# Adjacency-Faithfulness and Conservative Causal Inference

**Joseph Ramsey    Peter Spirtes**
Department of Philosophy
Carnegie Mellon University
Pittsburgh, PA 15213
{jdramsey, ps7z}@andrew.cmu.edu

**Jiji Zhang**
Division of Humanities and Social Sciences
California Institute of Technology
Pasadena, CA 91125
jiji@hss.caltech.edu

## Abstract

Most causal discovery algorithms in the literature exploit an assumption usually referred to as the Causal Faithfulness or Stability Condition. In this paper, we highlight two components of the condition used in constraint-based algorithms, which we call "Adjacency-Faithfulness" and "Orientation-Faithfulness." We point out that assuming Adjacency-Faithfulness is true, it is possible to test the validity of Orientation-Faithfulness. Motivated by this observation, we explore the consequence of making only the Adjacency-Faithfulness assumption. We show that the familiar PC algorithm has to be modified to be correct under the weaker, Adjacency-Faithfulness assumption. The modified algorithm, called Conservative PC (CPC), checks whether Orientation-Faithfulness holds in the orientation phase, and if not, avoids drawing certain causal conclusions the PC algorithm would draw. However, if the stronger, standard causal Faithfulness condition actually obtains, the CPC algorithm outputs the same pattern as the PC algorithm does in the large sample limit.

We also present a simulation study showing that the CPC algorithm runs almost as fast as the PC algorithm, and outputs significantly fewer false causal arrowheads than the PC algorithm does on realistic sample sizes.

## 1 MOTIVATION: FAITHFULNESS DECOMPOSED

Directed acyclic graphs (DAGs) can be interpreted both probabilistically and causally. Under the causal interpretation, a DAG $G$ represents a causal structure such that $A$ is a direct cause of $B$ just in case there is a directed edge from $A$ to $B$ in $G$. Under the probabilistic interpretation, a DAG $G$, also referred to as a Bayesian network, represents a probability distribution $P$ that satisfies the *Markov Property*: each variable in $G$ is independent of its non-descendants conditional on its parents. The Causal Markov Condition is a bridge principle linking the causal interpretation of a DAG to the probabilistic interpretation.[1]

**Causal Markov Condition**: Given a set of variables whose causal structure can be represented by a DAG $G$, every variable is probabilistically independent of its non-effects (non-descendants in $G$) conditional on its direct causes (parents in $G$).

The assumption that the causal structure can be represented by a DAG entails that there is no causal feedback, and that no common cause of any pair of variables in the DAG is left out. All DAG-based causal discovery algorithms assume the causal Markov condition, and most of them (e.g., those discussed in Pearl 2000, Spirtes et al. 2000, Heckerman et al. 1999) also assume, if only implicitly, the converse principle, known as the Causal Faithfulness or Stability Condition:

**Causal Faithfulness Condition**: Given a set of variables whose causal structure can be represented by a DAG, no conditional independence holds unless entailed by the Causal Markov Condition.

Conditional independence relations entailed by the Markov condition are captured exactly by a graphical criterion called d-separation (Neapolitan 2004), defined as follows. Given a path $p$ in a DAG, a non-endpoint vertex $V$ on $p$ is called a *collider* if the two edges incident to $V$ on $p$ are both into $V$ ($\rightarrow V \leftarrow$), otherwise $V$ is called a *non-collider* on $p$.

**Definition 1 (d-separation).** *In a DAG, a path $p$ between vertices $A$ and $B$ is* **active** *(d-connecting) relative to a set of vertices* **C** *($A, B \notin$ **C***) if*

---
[1]For a more formal presentation of the notions mentioned in this section, see Spirtes et al. (2000).

i. every non-collider on p is not a member of **C**;

ii. every collider on p is an ancestor of some member of **C**.

Two sets of variables **A** and **B** are said to be **d-separated** by **C** if there is no active path between any member of **A** and any member of **B** relative to **C**.

A well-known important result is that for any three disjoint sets of variables **A**, **B** and **C** in a DAG $G$, **A** and **B** are entailed (by the Markov condition) to be independent conditional on **C** if and only if they are d-separated by **C** in $G$. So the causal Faithfulness condition can be rephrased as saying that *for every three disjoint sets of variables* **A**, **B** *and* **C**, *if* **A** *and* **B** *are not d-separated by* **C** *in the causal DAG, then* **A** *and* **B** *are not independent conditional on* **C**.

Two simple facts about d-separation are particularly relevant to our purpose (see e.g. Neapolitan 2004, pp. 89 for proofs):

**Proposition 1.** *Two variables are adjacent in a DAG if and only if they are not d-separated by any subset of other variables in the DAG.*

Call a triple of variables $\langle X, Y, Z \rangle$ in a DAG an *unshielded triple* if $X$ and $Z$ are both adjacent to $Y$ but are not adjacent to each other.

**Proposition 2.** *In a DAG, any unshielded triple $\langle X, Y, Z \rangle$ is a collider if and only if all sets that d-separate $X$ from $Z$ do not contain $Y$; it is a non-collider if and only if all sets that d-separate $X$ from $Z$ contain $Y$.*

Below we focus on two implications of the Causal Faithfulness Condition, easily derivable given Propositions 1 and 2. We call them *Adjacency-Faithfulness* and *Orientation-Faithfulness*, respectively.

**Implication 1 (Adjacency-Faithfulness).** *Given a set of variables* **V** *whose causal structure can be represented by a DAG $G$, if two variables $X, Y$ are adjacent in $G$, then they are dependent conditional on any subset of* $\mathbf{V} \setminus \{X, Y\}$.

We call this condition Adjacency-Faithfulness for the obvious reason that this is the part of the Faithfulness condition that is used to justify the step of recovering adjacencies in constraint-based algorithms. Generically, this step proceeds by searching for a conditioning set that renders two variables independent, and by the causal Markov and Adjacency-Faithfulness conditions, the two variables are not adjacent if and only if such a conditioning set is found.

**Implication 2 (Orientation-Faithfulness).** *Given a set of variables* **V** *whose causal structure can be represented by a DAG $G$, let $\langle X, Y, Z \rangle$ be any unshielded triple in $G$.*

(O1) *if $X \to Y \leftarrow Z$, then $X$ and $Z$ are dependent given any subset of $\mathbf{V} \setminus \{X, Z\}$ that contains $Y$;*

(O2) *otherwise, $X$ and $Z$ are dependent conditional on any subset of $\mathbf{V} \setminus \{X, Z\}$ that does not contain $Y$.*

Orientation-Faithfulness obviously serves to justify the step of identifying *unshielded colliders* (and *unshielded non-colliders*). For any unshielded triple $\langle X, Y, Z \rangle$ resulting from the adjacency step, a conditioning set that renders $X$ and $Z$ independent must have been found. The Orientation-Faithfulness condition then implies that the triple is an unshielded collider if and only if the conditioning set does not contain $Y$. This is in fact what the familiar PC algorithm checks. The rest of our paper is motivated by the following simple observation: assuming the Adjacency-Faithfulness condition is true, we can in principle test whether Orientation-Faithfulness fails of a particular unshielded triple. Suppose we have a perfect oracle of conditional independence relations, which is in principle available for many parametric families in the large sample limit by performing statistical tests. Since the Adjacency-Faithfulness is by assumption true, out of the oracle one can construct correct adjacencies and non-adjacencies, and thus correct unshielded triples in the causal graph. For such an unshielded triple, say, $\langle X, Y, Z \rangle$, if there is a subset of $\mathbf{V} \setminus \{X, Z\}$ containing $Y$ that renders $X$ and $Z$ independent and a subset not containing $Y$ that renders $X$ and $Z$ independent, then Orientation-Faithfulness fails on this triple. This failing condition can of course be verified by the oracle.

Note that this simple test of Orientation-Faithfulness does not rely on knowing what the true causal DAG is. The reason why this test works is that a distribution that satisfies the Adjacency-Faithfulness with respect to the true causal DAG but fails the above test is not Orientation-Faithful to *any* DAG, and hence not Orientation-Faithful to the true causal DAG.

This suggests that theoretically we can relax the standard causal Faithfulness assumption and still have provably correct and informative causal discovery procedures. In fact, one main result we will establish in this paper is that the PC algorithm, though incorrect under the weaker, Adjacency-Faithfulness condition, can be revised in such a way that the modified version – that we call CPC (conservative PC) – is correct given the Adjacency-Faithfulness condition, and is as informative as the standard PC algorithm if the Causal Faithfulness Condition actually obtains.

In addition to the theoretical demonstration, we will present a simulation study comparing the CPC algo-rithm and the PC algorithm. The results show that the CPC algorithm runs almost as fast as the PC algorithm, which is known for its computational feasibility. More importantly, even when the standard Causal Faithfulness Condition holds, the CPC algorithm turns out to be more accurate on realistic sample sizes than the PC algorithm in that it outputs significantly fewer false causal arrowheads and (almost) as many true causal arrowheads.

## 2 HOW PC ALGORITHM ERRS

Before we present our modification of the PC algorithm, it is helpful to explain how the PC algorithm can make mistakes under the causal Markov and Adjacency-Faithfulness conditions. The relevant details of the PC algorithm are reproduced below, where we use $ADJ(G, X)$ to denote the set of nodes adjacent to $X$ in a graph $G$:

**PC Algorithm**

S1 Form the complete undirected graph $U$ on the set of variables $\mathbf{V}$;

S2 $n = 0$
   **repeat**

   For each pair of variables $X$ and $Y$ that are adjacent in (the current) $U$ such that $ADJ(U, X)\backslash\{Y\}$ or $ADJ(U, Y)\backslash\{X\}$ has at least $n$ elements, check through the subsets of $ADJ(U, X)\backslash\{Y\}$ and the subsets of $ADJ(U, Y)\backslash\{X\}$ that have exactly $n$ variables. If a subset $S$ is found conditional on which $X$ and $Y$ are independent, remove the edge between $X$ and $Y$ in $U$, and record $S$ as $Sepset(X, Y)$;

   $n = n + 1$;

   **until** for each ordered pair of adjacent variables $X$ and $Y$, $ADJ(U, X)\backslash\{Y\}$ has less than $n$ elements.

S3 Let $P$ be the graph resulting from step S2. For each unshielded triple $\langle A, B, C \rangle$ in $P$, orient it as $A \rightarrow B \leftarrow C$ iff $B$ is not in $Sepset(A, C)$.

S4 Execute the orientation rules given in Meek 1995.[2]

If the input to the PC algorithm is a sample from a population distribution that is faithful to some DAG, then in the large sample limit, the output of the PC algorithm can be interpreted as a set of DAGs, all of which are d-separation equivalent (that is, they imply exactly the same d-separation relations). The d-separation equivalence class of DAGs output by the PC algorithm (and the score-based GES algorithm as well) is represented by a graphical object called a pattern, or a PDAG (Chickering 2002). A pattern is a mixture of directed and undirected edges. A DAG is represented by a pattern if it contains the same adjacencies as the pattern, every directed edge $A \rightarrow B$ in the pattern is oriented as $A \rightarrow B$ in the DAG, and if the DAG contains an unshielded collider, then so does the pattern. The output of the PC algorithm is correct given the Causal Markov and Faithfulness Conditions and a perfect conditional independence oracle (such as statistical tests in the large sample limit) in the sense that the true causal DAG is among the DAGs represented by the output pattern. The output of the PC algorithm is complete in the sense that if an edge $A \rightarrow B$ occurs in every DAG in the d-separation equivalence class represented by the output pattern, then it is oriented as $A \rightarrow B$ in the output pattern. (Meek 1995, Spirtes et al. 2000).

Two specific features of PC are worth noting. First, in S2, the adjacency step, the PC algorithm essentially searches for a conditioning set for each pair of variables that renders them independent, which we henceforth call a *screen-off* conditioning set. But it does this with two additional tricks: (1) it starts with the conditioning set of size 0 (i.e., the empty set) and gradually increases the size of the conditioning set; and (2) it confines the search of a screen-off conditioning set for two variables to within the potential parents – i.e., the currently adjacent nodes – of the two variables, and thus systematically narrows down the space of possible screen-off sets as the search goes on. These two tricks increase both computational and statistical efficiency in most real cases, and we will keep this step intact in our modification.

Secondly, in S3 the PC algorithm uses a very simple rule to identify unshielded colliders or non-colliders. For any unshielded triple $\langle X, Y, Z \rangle$, it simply checks whether or not $Y$ is contained in the screen-off set for $X$ and $Z$ found in the adjacency stage. Now if we assume the causal Markov and Adjacency-Faithfulness conditions are true, the adjacencies (and non-adjacencies) resulting from the adjacency stage are asymptotically correct. However, these two conditions do not imply the truth of Orientation-Faithfulness, and when the latter fails, the PC algorithm will err even in the large sample limit.

Consider the simplest example $A \rightarrow B \rightarrow C$ where $A \perp\!\!\!\perp C$ and $A \perp\!\!\!\perp C | B$. This is the case when, for example, causation fails to be transitive, an issue of great interest to philosophers of causality. In this situation

---
[2] Details of the Meek orientation rules do not matter for the purposes of this paper. The rules are also described in Neapolitan 2004, pp. 542.

the causal Markov and Adjacency-Faithfulness conditions are both satisfied, but Orientation-Faithfulness is not true of the triple $\langle A, B, C \rangle$. Now, given the correct conditional independence oracle, the PC algorithm would remove the edge between $A$ and $C$ in S2 because $A \perp\!\!\!\perp C$, and later in S3 orient the triple as $A \rightarrow B \leftarrow C$ because $B$ is not in the screen-off set found in S2, i.e., the empty set. Simple as it is, the example suffices to establish that the PC algorithm is not asymptotically correct[3] under the causal Markov and Adjacency-Faithfulness assumptions.

## 3   CONSERVATIVE PC

It is not hard, however, to modify the PC algorithm to retain correctness under the weaker assumption. Indeed a predecessor of the PC algorithm, called the SGS algorithm (Spirtes et al. 2000), is almost correct. The SGS algorithm decides whether an unshielded triple $\langle X, Y, Z \rangle$ is a collider or a non-collider by literally checking whether (O1) or (O2) in the statement of Orientation-Faithfulness is true. Theoretically all it lacks is a clause that acknowledges the failure of Orientation-Faithfulness when neither (O1) nor (O2) passes the check. Practically, however, the SGS algorithm is a terribly inefficient algorithm. Computationally, it is best case exponential because it has to check dependence between $X$ and $Z$ conditional on every subset of $\mathbf{V} \backslash \{X, Z\}$. Statistically, tests of independence conditional on large sets of variables have very low power, and are likely to lead to errors. In addition, the sheer number of conditional independence tests makes it exceedingly likely that some of them will err, and we suspect that almost every unshielded triple will be marked as unfaithful if we run the SGS algorithms on more than a few variables.

Fortunately, the main idea of the PC algorithm comes to the rescue. A correct algorithm does not have to check every subset of $\mathbf{V} \backslash \{X, Z\}$ in order to test whether $\langle X, Y, Z \rangle$ is a collider, a non-collider, or an unfaithful triple. It only needs to check subsets of the variables that are potential parents of $X$ and $Z$. This trick, as we shall show shortly, is theoretically valid, and turns out to work well in simulations.

The CPC algorithm replaces S3 in PC with the following S3', and otherwise remains the same.

S3' Let $P$ be the graph resulting from step 1. For each unshielded triple $\langle A, B, C \rangle$, check all subsets of $A$'s potential parents and of $C$'s potential parents:

(a) If $B$ is NOT in any such set conditional on which $A$ and $C$ are independent, orient $A - B - C$ as $A \rightarrow B \leftarrow C$;
(b) if $B$ is in all such sets conditional on which $A$ and $C$ are independent, leave $A - B - C$ as it is, i.e., a non-collider;
(c) otherwise, mark the triple as "unfaithful" by underlining the triple, $A - \underline{B} - C$.

(In S4, The orientation rules that are applied to unshielded non-colliders in the PC algorithm are, of course, applied only to unshielded non-colliders in the CPC algorithm; in particular they are not applied to triples that are marked as unfaithful.)

The output of the CPC algorithm can also be interpreted as a set of DAGs. If the input to the CPC algorithm is a sample from a distribution that satisfies the Markov and Adjacency-Faithfulness Assumptions, in the large sample limit, the output is an extended pattern, or e-pattern for short. An e-pattern contains a mixture of undirected and directed edges, as well as underlinings for unshielded triples that are unfaithful. A DAG is represented by an e-pattern if it has the same adjacencies as the e-pattern, every directed edge $A \rightarrow B$ in the e-pattern is oriented as $A \rightarrow B$ in the DAG, and every unshielded collider in the DAG is either an unshielded collider or a marked unfaithful triple in the e-pattern. These rules allow that an unfaithful triple in the e-pattern can be oriented as either a collider or a non-collider in a DAG represented by the e-pattern.

The set of DAGs represented by an e-pattern may not be d-separation equivalent, if the e-pattern contains an unfaithful triple. For example, if $A$ causes $B$, and $B$ causes $C$, but the causation is not transitive (i.e. $I(A, C|B)$ and $I(A, C)$), the resulting e-pattern is $A - \underline{B} - C$, because it is an unfaithful triple. The set of DAGs represented by $A - \underline{B} - C$ contains $A \rightarrow B \rightarrow C$, $A \leftarrow B \rightarrow C$, $A \leftarrow B \leftarrow C$, and $A \rightarrow B \leftarrow C$. The latter DAG is not d-separation equivalent to the first three DAGs. Note that in this case the true distribution lies in the intersection of sets of distributions represented by non- d-separation equivalent DAGs. The intersection would be ruled out as impossible by the standard Faithfulness assumption.

At this point it should be clear why the modified PC algorithm is labeled "conservative": it is more cautious than the PC algorithm in drawing unambiguous conclusions about causal orientations. A typical output of the CPC algorithm is shown in Figure 1. The conservativeness is of course what is needed to make the algorithm correct under the causal Markov and Adjacency-Faithfulness assumptions.

---

[3] By "asymptotically correct" we mean the probability of the output containing an error converges to zero in the large sample limit, no matter what the true probability distribution is.

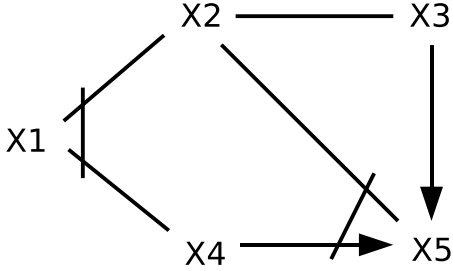

Figure 1: A typical output for CPC. Underlining (which in the figure looks like "crossing") indicates unfaithful unshielded triples discovered by the algorithm.

**Theorem 1 (Correctness of CPC).** *Under the causal Markov and Adjacency-Faithfulness assumptions, the CPC algorithm is correct in the sense that given a perfect conditional independence oracle, the algorithm returns an e-pattern that represents the true causal DAG.*

*Proof.* Suppose the true causal graph is $G$, and all conditional independence judgments are correct. The Markov and Adjacency-Faithfulness assumptions imply that the undirected graph $P$ resulting from step $S2$ has the same adjacencies as $G$ does (Spirtes et al. 2000). Now consider step $S3'$. If $S3'(a)$ obtains, then $A \to B \leftarrow C$ must be a subgraph of $G$, because otherwise by the Markov assumption, either $A$'s parents or $C$'s parents d-separate $A$ and $C$, which means that there is a subset $\mathbf{S}$ of either $A$'s potential parents or $C$'s potential parents containing $B$ such that $A \perp\!\!\!\perp C | \mathbf{S}$, contradicting the antecedent in $S3'(a)$. If $S3'(b)$ obtains, then $A \to B \leftarrow C$ cannot be a subgraph of $G$ (and hence the triple must be an unshielded non-collider), because otherwise by the Markov assumption, there must be a subset $\mathbf{S}$ of either $A$'s potential parents or $C$'s potential parents not containing $B$ such that $A \perp\!\!\!\perp C | \mathbf{S}$, contradicting the antecedent in $S3'(b)$. So neither $S3'(a)$ nor $S3'(b)$ will introduce an orientation error. It follows that every unshielded collider in $G$ is either an unshielded collider or a marked triple in $P$. Trivially $S3'(c)$ does not produce an orientation error, and it has been proven (in e.g., Meek 1995) that $S4$ will not produce any, which implies that every directed edge in $P$ is also in $G$. □

The theorem entails that the output e-pattern (1) has the same adjacencies as the true causal DAG; and (2) all arrowheads and unshielded non-colliders in the e-pattern are also in the true causal DAG. Theorem 1, together with the consistency of statistical tests of in-dependence, entails that the probability of the output containing an error approaches zero as the sample size approaches infinity.

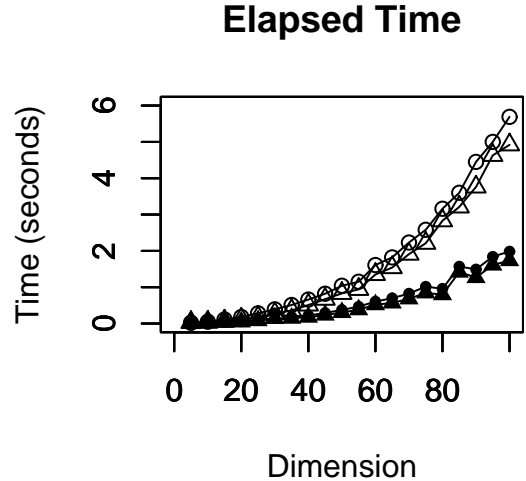

Figure 2: Average elapsed time

Note that a triple $A \underline{\to B \leftarrow} C$ or $A \underline{-- B \to} C$ may occur in cases where the triple was initially marked as unfaithful, but all later orientation rules provided further consistent orientation information. In those cases, the underlining serves no purpose (as ambiguity concerning collider vs. non-collider is already dissolved) and can be removed. The remaining triples marked unfaithful by the CPC algorithm in the large sample limit are truly ambiguous in that either a collider or a non-collider is compatible with the conditional independence judgments. We conjecture but cannot yet prove that the CPC algorithm is complete in the sense that for every undirected edge in an e-pattern output by the CPC algorithm, there is a DAG represented by the e-pattern that orients the edge in one direction and another DAG represented by the e-pattern that orients the edge in the other direction. Finally, it is obvious that asymptotically the CPC algorithm and the PC algorithm produce the same output if the standard Faithfulness assumption actually obtains.

## 4 SIMULATION RESULTS

The theoretical superiority of the CPC algorithm over the PC algorithm may not necessarily cash out in practice if the situations where the Adjacency-Faithfulness but not the Orientation-Faithfulness holds do not arise often. We will not try to make an argument to the contrary here, even though we believe such an argu-ment can be made. Instead we wish to show that the CPC algorithm in practice performs better than the PC algorithm, regardless of whether Orientation-Faithfulness holds or not. That is, even when the data are generated from a distribution Markov and Faithful to the true causal graph, it pays to be conservative on realistic sample sizes. One possible rationale for this is that even though PC is correct in the large sample limit if Orientation-Faithfulness is not violated, it is very liable to error on realistic sample sizes if Orientation-Unfaithfulness is almost violated. Almost violations of Orientation-Faithfulness can arise in several ways – for example, when a triple chain is almost non-transitive, or more generally, when one of the edges in an unshielded triple is very weak[4].

The simulations illustrate that the extra independence checks invoked in the CPC algorithms do not render CPC significantly slower than PC and that CPC is more accurate than PC in terms of arrow orientations. The explanation for the first point is that the main computational expense of the PC algorithm occurs in the adjacency stage; the number of independence checks added in CPC for orientation is small by comparison. The second point suggests that the PC algorithm too often infers that unshielded triples are colliders, and the CPC algorithm provides the right antidote to this by means of the extra checks it performs. Again, we expect that the CPC algorithm will do particularly better than the PC algorithm when the distribution generated is close-to-unfaithful to the true graph – a situation pointed out by several authors as a major obstacle to reliable causal inference (Robins et al. 2003, Zhang and Spirtes 2003).

To illustrate these points, the following simulations were performed on linear Gaussian models, with variations for sparser and denser graphs, with dimensions (numbers of variables) ranging from 5 to 100 variables. For the sparser case, for each dimension $d$ from 5 to 100 in increments of 5, five random graphs were selected uniformly from the space of DAGs with at most $d$ edges and with a maximum degree of 10. For each such graph, a random structural equation model was constructed by selecting edge coefficients randomly 0.95 of the time uniformly from $[-1.5, -0.5] \cup [0.5, 1.5]$ and 0.05 of the time uniformly from $[-0.001, 0.001]$. (Selection from the range $[-0.001, 0.001]$ guarantees the presence of weak edges, which in turn often lead to al-

---

[4]Intuitively, almost violations of Orientation-Faithfulness refer to situations where two variables, though entailed to be dependent conditional on some variables by the Orientation-Faithfulness condition, are close to conditionally independent. How to quantify the "closeness" and just how close is close enough to cause trouble depend on distributional assumptions and sample sizes.

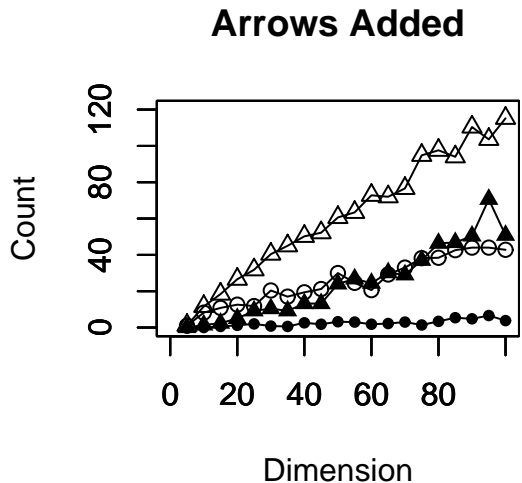

**Arrows Added**

Figure 3: Average count of arrow false positives

most violations of faithfulness.) For each such model, a random data set of 1000 cases was simulated, to which PC and CPC were applied with significance level $\alpha = 0.05$ for each hypothesis test of conditional independence, tested using Fisher's Z transformation of partial correlation. The output was compared to the pattern of the true DAG (the true pattern), not the true DAG itself. Performance statistics were recorded, including elapsed time and false positive and negative counts for arrows, unshielded colliders, unshielded non-colliders, and adjacencies. For each number of variables, each performance statistic was averaged over the five random models constructed at that dimension, for PC and for CPC. This procedure was repeated for denser models with DAGs randomly selected uniformly from the set of DAGs with at most $2d$ edges and a maximum degree of 10.

Counting orientation errors when there are differences in adjacencies as well raises some subtle issues that we have chosen to resolve in the following way. An arrowhead removal error (false negative) occurs when the true pattern $P_1$ contains $A \rightarrow B$, but the output $P_2$ either does not contain an edge between $A$ and $B$ or does contain an edge between $A$ and $B$ but there is no arrowhead on this edge at $B$. An analogous rule is used to count arrowhead addition errors (false positive). This has the consequence that if $A$ and $B$ are not adjacent in $P_1$, but $A - B$ is in $P_2$, this is counted as an adjacency addition error, but not an arrowhead addition or removal error. In contrast, if $A \rightarrow B$ is in $P_2$, this is counted as an adjacency addition error and an arrowhead addition error, because of the arrowhead at $B$. The justification for this is that the $A - B$ er-

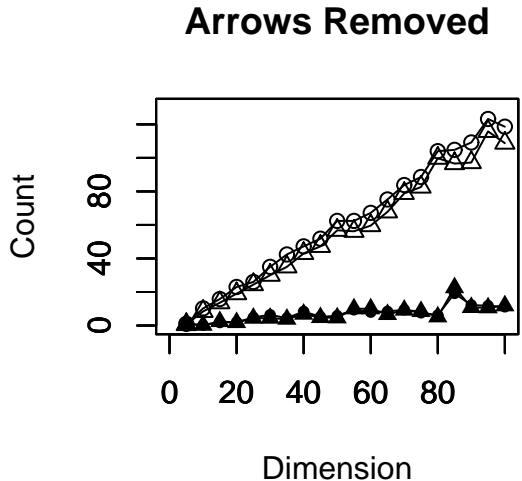

**Arrows Removed**

Figure 4: Average count of arrow false negatives

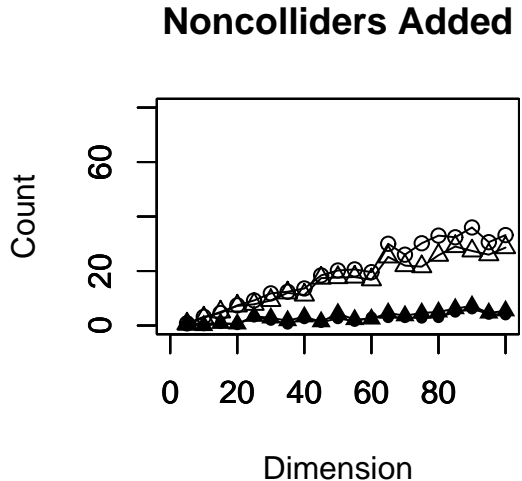

**Noncolliders Added**

Figure 5: Average count of non-collider false positive

ror leaves open whether there is an arrowhead at $B$, and does not lead to any errors in predicting the effects of manipulations (the effects of manipulation are unknown because the orientation of the edge is unknown). In contrast, the $A \rightarrow B$ error does definitively state that there is an arrowhead at $B$, and does lead to errors in predicting the effects of manipulations.

There is an unshielded non-collider addition error for the triple $\langle X, Y, Z \rangle$ if they form an unshielded non-collider in $P_2$, but $P_1$ either has different adjacencies among $X$, $Y$, and $Z$, or the same adjacencies but is a collider. An unfaithful triple in $G_2$ does not count as an unshielded non-collider or collider addition error, regardless of what is in $G_1$. Unshielded non-collider removal errors are handled in an analogous fashion.

In all the figures, PC statistics are represented by triangles and CPC statistics are represented by circles; sparser models use filled symbols, and denser models used unfilled symbols. The horizontal axis is the number of variables in the true DAG.

Figure 2 shows that for both sparser and denser models, CPC is only slightly slower than PC.

Figure 3 shows that for both sparser and denser models, the number of extra arrows introduced is far better controlled by CPC than by PC. For sparser models, the number is particularly well-controlled. Figure 4 shows that for both sparser and denser models, the number of arrowhead removal errors committed by CPC is almost indistinguishable from the number of arrowhead removal errors committed by PC.

The performance of CPC regarding false positive and false negative unshielded non-colliders also matches that of PC, as shown in Figures 5 and 6. In a word, in no respect is PC noticeably better than CPC, whereas CPC is significantly better than PC in avoiding false positive causal arrowheads, the arguably most consequential type of errors.

Figure 7 plots the percentage of unfaithful triples among the total number of unshielded triples output by CPC. For sparser models, the percentage of unfaithful triples is around 30 percent; for denser models, it rises to around 40 percent. This confirms our expectation that CPC is more conservative the denser the true graph.

Similar simulations were carried out parameterizing random graphs using discrete Bayes nets with 2 to 4 categories per variable, but otherwise with identical setup to the sparser continuous simulations above, with similar results.

## 5 CONCLUSION

The CPC algorithm we proposed in this paper is provably correct under the causal Markov assumption plus a weaker-than-standard Faithfulness assumption, the Adjacency-Faithfulness assumption. It can be regarded as a conservative generalization of the PC algorithm in that it theoretically gives the same answer as the PC does under the standard assumptions. Perhaps more importantly, simulation results suggest that the CPC algorithm works much better than the PC algorithm in terms of avoiding false causal arrowheads, and achieves this without costing significantly more in running time or missing positive information. We do not claim that the evidence is conclusive, and we think it would be interesting to compare CPC and PC on real data sets.

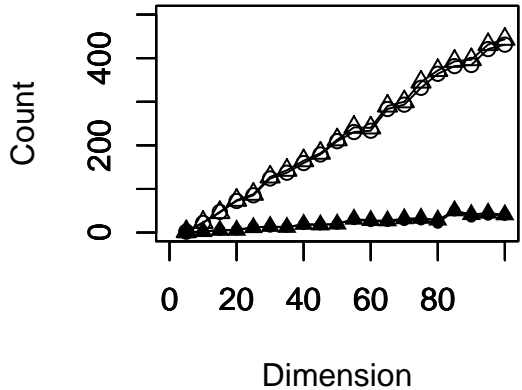

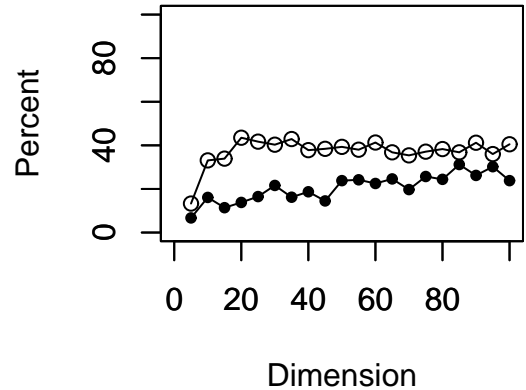

Figure 6: Average count of non-collider false negatives

Figure 7: Percent of unfaithful triples among all unshielded triples

A natural question is how a score-based search algorithm would perform when the true distribution is only Adjacency-Faithful to the true causal graph. Such algorithms could err even in the large sample limit given typical scores such as BIC. For example, for three discrete variables, if the true causal graph is $A \rightarrow B \leftarrow C$, but the true distribution satisfies $A \perp\!\!\!\perp C$ and $A \perp\!\!\!\perp C | B$, which violates the Orientation-Faithfulness condition, the GES algorithm (Chickering 2002) outputs $A \text{---} B \text{---} C$, an unshielded non-collider, given large sample. This is because such scores as BIC would (eventually) prefer the model with fewer parameters when both contain the true distribution.

Another possible twist to the CPC algorithm is that when a sign of an unfaithful triple arises, it actually suggests that extra checks on the adjacencies in the triple should be performed. We are currently exploring this and other ideas that can potentially increase the accuracy of the estimated adjacencies. Another ongoing project is to extend our work in this paper to the FCI algorithm, which, unlike the PC algorithm, does not make the often unrealistic assumption of no unmeasured common causes.

### Acknowledgements

We thank anonymous referees for valuable comments.